\documentclass[letterpaper]{article} 
\usepackage[]{aaai25}  
\usepackage{times}  
\usepackage{helvet}  
\usepackage{courier}  
\usepackage[hyphens]{url}  
\usepackage{graphicx} 
\usepackage{natbib}  
\usepackage{caption} 
\usepackage{algorithm}
\usepackage{algorithmic}

\usepackage[table]{xcolor}
\usepackage{booktabs}
\usepackage{amsmath}
\usepackage{amsfonts}

\newcommand{\myfigref}[1]{Figure \ref{#1}}
\newcommand{\mytabref}[1]{Table \ref{#1}}
\usepackage{multirow} 

\usepackage{xcolor}
\definecolor{green1}{HTML}{0aa344}
\definecolor{green2}{HTML}{01b83a}
\definecolor{orange}{HTML}{fe793d}
 
\usepackage{newfloat}
\usepackage{listings}
\DeclareCaptionStyle{ruled}{labelfont=normalfont,labelsep=colon,strut=off} 
\floatstyle{ruled}
\newfloat{listing}{tb}{lst}{}
\floatname{listing}{Listing}
\title{StylePrompter: Enhancing Domain Generalization with Test-Time Style Priors}
\author {
    Jiao Zhang\textsuperscript{\rm 1\rm 2},
    Jian Xu\textsuperscript{\rm 1\rm 2},
    Xu-Yao Zhang\textsuperscript{\rm 1\rm 2}
    Cheng-Lin Liu\textsuperscript{\rm 1\rm 2}
}
\affiliations {
    \textsuperscript{\rm 1}MAIS, Institute of Automation, Chinese Academy of Sciences, China\\
    \textsuperscript{\rm 2}School of Artificial Intelligence, University of Chinese Academy of Sciences, China\\
    \{zhangjiao2019, jian.xu\}@ia.ac.cn, \{xyz, liucl\}@nlpr.ia.ac.cn
}

\usepackage{bibentry}

\begin{document}

\maketitle

\begin{abstract}
  In real-world applications, the sample distribution at the inference stage often differs from the one at the training stage, causing performance degradation of trained deep models. The research on domain generalization (DG) aims to develop robust algorithms that can improve the generalized performance in unseen domains by training on a few domains. However, the domain-agnostic vision model, trained on a limited number of domains using traditional domain generalization methods, cannot guarantee its effectiveness in dealing with unseen domains. The introduction of language can break the closed cognition space of the vision model, providing additional semantic information that cannot be inferred from vision-only datasets. In this paper, we propose to overcome the challenge in previous DG methods by introducing the style prompt in the language modality to adapt the trained model dynamically. In particular, we train a style prompter to extract style information of the current image into an embedding in the token embedding space and place it in front of the candidate category words as prior knowledge to prompt the model. Our open space partition of the style token embedding space and the hand-crafted style regularization enable the trained style prompter to handle data from unknown domains effectively. Extensive experiments verify the effectiveness of our method and demonstrate state-of-the-art performances on multiple public datasets. Codes will be available after the acceptance of this paper.
\end{abstract}

\section{Introduction}
Deep neural networks have achieved superior performance on many in-laboratory datasets, but the trained models are prone to suffer performance degradation in real-world applications, which is caused by two main reasons. First, the popular end-to-end training paradigm makes the deep model have poor interpretability and tends to learn a statistical bias toward the training data. Second, the distribution of test data often deviates from training data due to unpredictable environmental changes and the bias introduced in the data collection process. The above problems cause deep models with huge risks and costs in some high-stakes open-world applications, such as intelligent entrance guards, driverless cars, and disease diagnosis. Therefore, how deep models can maintain stable performance on unseen out-of-distribution data is a critical problem to be solved.

\begin{figure}[t]
  \centering
  \includegraphics[width=1\linewidth]{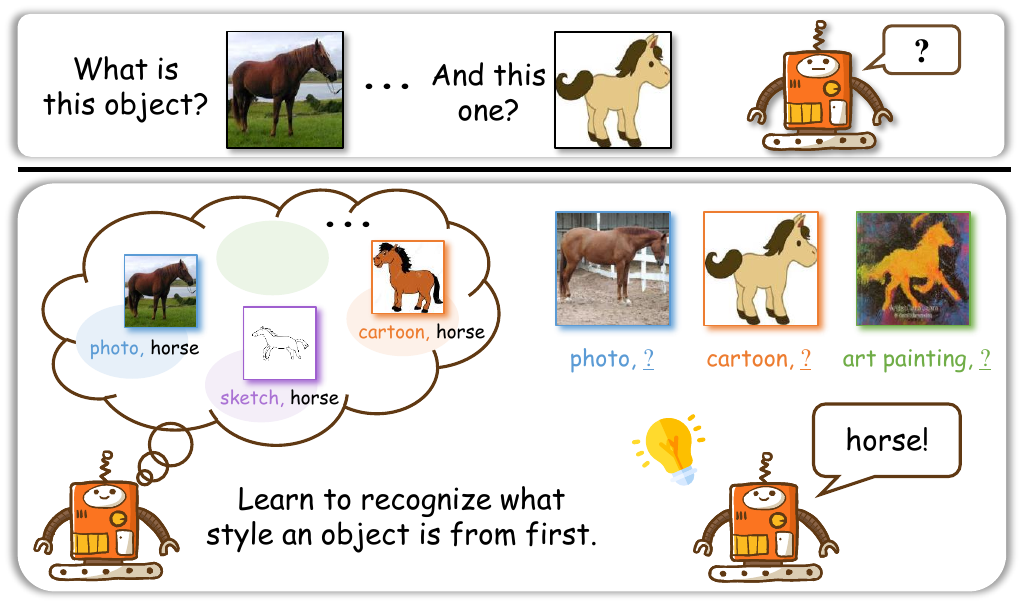}
  \caption{Recognizing objects from unseen domains is a challenging task for vision models. In this paper, we enhance the model's domain generalization ability by providing test-time style priors as prompts in the language modality.}
  \label{motivation}
  \vskip -0.1in
\end{figure}

In response to the problem of distribution shifts encountered in real-world applications, domain generalization (DG) methods have received increasing attention to improving the deep model's generalization ability on unknown domains. In the past decade, numerous studies have been made from different perspectives, which can be roughly categorized as data manipulation, representation learning, and learning strategy \cite{WangLLOQLCZY23}. Methods based on data manipulation \cite{VolpiNSDMS18, ZhouY0X21} study to increase the diversity of training data and expand the distribution of original training data, thereby improving the model's generalization ability for unknown domain data. Representation learning methods aim to learn robust feature representations to cope with sample variations, including domain-invariant representation learning methods \cite{ErfaniBMNLBR16, LiGTLT18} and feature disentanglement methods \cite{PengHSS19, NamLPYY21}. In addition, many methods attack the DG problem from the perspective of model optimization. These methods improve the model's generalization ability by applying universal robust machine learning algorithms or heuristic training strategies, such as ensemble learning \cite{SeoSKKHH20}, meta-learning \cite{LiYSH18, LiYZH19}, self-supervised learning \cite{CarlucciDBCT19}, and gradient operations \cite{ParascandoloNOG21}. 

Despite the numerous progress in DG study, there is still no clear answer to what the most effective solution is. Two impressive benchmark works \cite{GulrajaniL21, KohSMXZBHYPGLDS21} also challenge the progress of domain generalization research. In the work \cite{GulrajaniL21}, the authors implemented fourteen methods on seven datasets under the same experimental conditions. Their results show that no method simultaneously outperforms ERM (empirical risk minimization) on all datasets, and no algorithm outperforms ERM by more than one percent in average performance. The work \cite{KohSMXZBHYPGLDS21} revealed that the out-of-distribution performance of the model trained by ERM is much lower than the in-distribution performance, and this gap still exists even with models trained by existing DG methods designed for tackling distribution shifts. The difficulty in current research on DG is that it is hard to learn a vision model that can cope with any unknown distribution shifts using a few training domains. Since the vision-only dataset implies a closed domain-variant space, there is no guarantee that the domain-invariant vision model learned on the training domains can generalize to unknown domains. Compared to vision, language provides additional semantic information that cannot be inferred from vision-only datasets. As shown in \myfigref{motivation}, we can break the closed style cognition space of the vision model by providing dynamic style prompts in the language modality, thus facilitating reliable domain generalization.

In this paper, we propose to improve the out-of-domain generalization performance of trained models by dynamically prompting the style information of input images with the assistance of pre-trained vision-language models. Specifically, we train a style prompter to extract the style information of input images, which takes image features as input and outputs the style information in the token embedding space. Our open partition of the style token embedding space and hand-crafted style regularization makes the style prompter extract correct style embedding for samples from unknown domains by only training on a few domains. During inference, we use the style information extracted by the proposed style prompter as prior knowledge of the current image and put it in front of the candidate category words to prompt the model. Thanks to the image-text matching ability of vision-language models, the style priors extracted from test images can further explore the capabilities of models pre-trained on large-scale image-text data and help the model adaptively deal with unknown domains. The experimental results validate the effectiveness of our method and show state-of-the-art results on four public datasets with different sizes.

In summary, our main contributions are as follows:
\begin{itemize}
  \item We propose to overcome the deficiency of traditional DG methods in extending the domain-invariant model learned from source domains to unknown domains by introducing style priors from input images to dynamically prompt the vision-language model at test time.
  \item The proposed open partition of the style token embedding space and hand-crafted style regularization enable our style prompter to correctly extract the style information of samples from unknown domains by only training on a few domains.
  \item Our method provides new insights and perspectives for research on domain generalization, improves the performance of DG methods on public datasets, and achieves state-of-the-art results.
\end{itemize}

\section{Related work}

\subsection{Domain Generalization}
Domain generalization aims to improve model performance on unseen domains by training on multiple source domains. Key approaches include \emph{domain alignment}, which aligns marginal or class-conditional distributions across source domains \cite{ErfaniBMNLBR16,LiPWK18,LiGTLT18}, and \emph{disentangled representations} \cite{PengHSS19,NamLPYY21}, which allow partial features to be domain-specific while others remain domain-agnostic. These methods typically require domain labels. Zhang et al. \cite{ZhangZWL23a} theoretically model DG using the information bottleneck principle and achieve excellent performance without requiring domain labels. Other methods focus on expanding data distributions through \emph{data augmentation} \cite{VolpiNSDMS18,ZhouYHX20,ZhouY0X21}, using \emph{meta-learning} to simulate domain generalization scenarios during training, boosting model performance by \emph{ensemble learning} \cite{SeoSKKHH20}, or designing \emph{regularization strategies} \cite{WangHLX19,HuangWXH20} to avoid learning false low-level features. Recently, due to the robustness of pre-trained vision-language models \cite{RadfordKHRGASAM21, JiaYXCPPLSLD21} to distribution shifts, some works \cite{RuanDM22,ChaLPC22,ChoNKYK23} have begun exploiting these models to develop new domain generalization methods. All these methods methods train models with a few domains and directly apply them without adaptation, making it difficult to handle any unknown distribution shifts. This paper proposes dynamically adapting the trained model using style information extracted from input images at test time, providing more reliable domain generalization.

\begin{figure*}[t]
  \centering
  \includegraphics[width=1\linewidth]{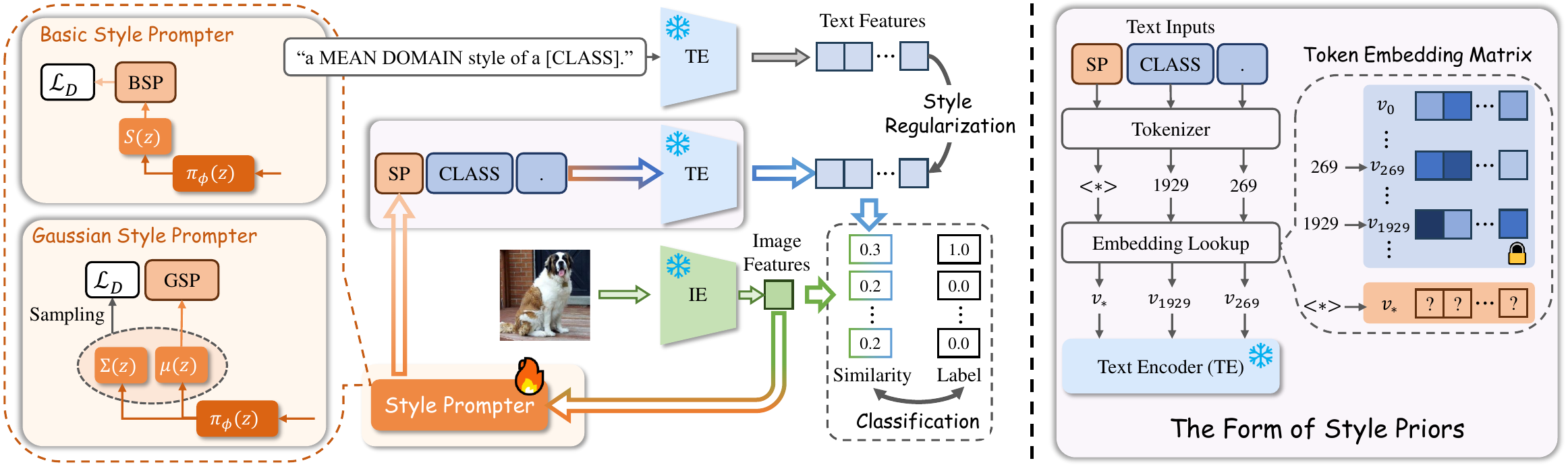}
  \caption{\textbf{Left} illustrates our method's framework, where we freeze the image and text encoders of a pre-trained vision-language model and train a lightweight style prompter (\textcolor{orange}{orange module}). Two designs are shown: the basic and Gaussian style prompters, where ``SP'', ``BSP'', and ``GSP'' represent ``style priors'', ``basic style priors'', and ``Gaussian style priors'', respectively. Contextualized style regularization is designed to enhance generalization and help integrate the learned style embeddings with category words. \textbf{Right} shows the form of style priors, which is dynamically updated based on the input image.}
  \label{framework_train}
  \vskip -0.1in
\end{figure*}

\subsection{Textual Inversion}
Textual inversion maps an input image to a pseudo-word for personalized image generation in text-to-image synthesis. First introduced in \cite{GalAAPBCC23}, it optimizes a pseudo-word embedding using 3-5 images to represent user-provided concepts, such as objects or styles, based on the reconstruction loss of a pre-trained text-to-image model. Kumari et al. \cite{KumariZ0SZ23} and Ruiz et al. \cite{RuizLJPRA23} extend this by simultaneously optimizing the pseudo-word embedding and fine-tuning the pre-trained models to create custom images based on a reference set. Daras et al. \cite{daras2022multiresolution} propose multiresolution textual inversion, which learns multiple pseudo-words to represent a concept at different resolutions, enabling personalized image generation with varying detail. Baldrati et al. \cite{BaldratiA0B23} apply textual inversion to composed image retrieval (CIR) by converting query images into pseudo-word token embeddings and concatenating them with query text. To speed up retrieval, they train a network for offline inference by distilling knowledge from pre-generated pseudo-word token embeddings. This paper applies textual inversion to domain generalization by extracting pseudo-word style embeddings from images to prompt the model, enhancing its ability to generalize to unknown domains.

\section{Method}
In this paper, we propose a novel and straightforward approach to enhance pre-trained vision-language models with improved domain generalization ability by extracting style information from input images as prior knowledge in the language modality. Our method, illustrated in \myfigref{framework_train}, dynamically adapts the model to the test environment in an offline manner. Unlike existing DG methods that apply the trained model to unknown domains without any adaptation, the proposed method offers better transferability. Compared to the test-time adaptation (TTA) method \cite{SunWLMEH20}, our method does not require updating model parameters at test time and thus has better practicality. The following sections provide a detailed introduction to the form of the proposed style priors, the network architecture and loss function for training the style prompter to extract the style priors, and the model inference process.

\subsection{Style Priors}
\label{Style_Priors}
Due to the limited training data, it is challenging to ensure that a model trained on a few source domains can handle any unknown distribution shifts. For unknown test samples, a more reliable approach is to first predict their style (distribution) information and use it as prior knowledge to assist model inference.

Thanks to the emergence of pre-trained vision-language models, which perform classification by matching test images with category words, we propose to help the model adapt to test samples from unknown domains using style prompts in the language modality. Specifically, we place the pseudo-word ``SP'', representing style information, in front of the category word ``CLASS'' and feed text in the form of ``SP [CLASS].'' into the language branch of the model. Before being sent to the encoder, the text inputs are preprocessed through the tokenizer and embedding lookup, as shown in right part of \myfigref{framework_train}. The tokenizer converts words into tokens, which are numerical representations, each corresponding to an entry in the model's dictionary. Based on the numerical representation of each token, we query the corresponding pre-trained token embedding in the token embedding matrix—this is the embedding lookup process. The proposed pseudo-word does not correspond to a pre-trained token embedding; instead, its embedding is dynamically extracted from the input image, condensing the style information of the current sample. After this process, the proposed style priors are fed into the text encoder together with the embeddings of the candidate category words, adapting the pre-trained model to the current environment.

In single-modal research, test-time adaptation (TTA) \cite{SunWLMEH20, WangSLOD21, ZhangLF22} also uses test data to dynamically adapt the model before making predictions. While these methods generally outperform DG methods, they are limited to scenarios where deployment devices have sufficient computing power and low real-time requirements to perform online learning. Compared to TTA, our inference process is offline, where the style information of test samples is extracted by the trained style prompter without the need to adapt the model.

\subsection{\textbf{Network Architecture}}
\label{Network_Architecture}
To obtain style priors in the language modality for test images, we designed a module for image-to-text conversion, termed style prompter (StylePrompter). Specifically, we use an image encoder and a text encoder as the backbone network, with their parameters initialized and frozen using a pre-trained vision-language model. We then train the style prompter to take image features as input and output the corresponding style priors, as shown in the left part of \myfigref{framework_train}. The structure of the style prompter is flexible. Two specific design schemes are detailed below, and their respective performances are presented in the following experiments.

\textbf{Basic Style Prompter.} To make image-to-text conversion, a straightforward approach is to use a shallow multi-layer perceptron (MLP) to connect the image feature space and the token embedding space, which performs space transformation and dimension alignment.

Specifically, as shown in the upper left of \myfigref{framework_train}, $\pi_\phi(\mathbf{z})$ is a manually designed network structure, and $S(\mathbf{z})$ is the style prompter model optimized at each step in the hyperparameter space of $\pi_\phi(\mathbf{z})$, where z is the features of the current input image. The $\pi_\phi(\mathbf{z})$ consists of two linear layers with the output size of 1/2 image feature dimension followed by the exponential linear unit (ELU) and a final linear layer with the output dimension the same as the token embedding dimension. This method trains a style prompter to extract a definitive style token embedding based on the input image. The embedding is used to calculate the domain discrimination loss $\mathcal{L}_{D}$ and perform the style prompt in the language branch of the model. While this approach ensures that the trained style prompter fits the training data well, we cannot guarantee that it maintains good style extraction performance on unseen test domains.

\textbf{Gaussian Style Prompter.} For Gaussian style prompter, we model the style token embedding space as a prior Gaussian distribution to prevent our style prompter from overfitting to the training domains. Specifically, we refer to the Gaussian modeling method by Derakhshani et al. \cite{DerakhshaniSBCS23}, which regularizes the prompt space, reduces overfitting to the seen classes, and improves generalization to unseen classes.

The Gaussian style prompter framework is shown in the bottom left of \myfigref{framework_train}. We define the style distribution of each input sample as a Gaussian distribution conditioned on the input image features \(\mathbf{z}\), represented as $\mathbf{s}(\mathbf{z}) \sim \mathcal{N}(\mu(\mathbf{z}), \Sigma(\mathbf{z}))$, where $\mu(\mathbf{z})$ and $\Sigma(\mathbf{z})$ are functional models parameterized based on the manually designed network structure $\pi_\phi(\mathbf{z})$. For domain discrimination, we use the reparameterization trick to generate Monte Carlo samples from the Gaussian distribution $\mathbf{s}(\mathbf{z})$. This process yields a total of $B\times N$ style samples, where \(B\) is the batch size and \(N\) is the number of Monte Carlo samples, which are used to train the Gaussian modeling style prompter. Since the mean $\mu(\mathbf{z})$ is the most informative with respect to the input image, we use it as the style prompt in front of the candidate category words and feed them into the text encoder (TE) to obtain the text embeddings of candidate categories.

\subsection{\textbf{Loss Function}}
\label{loss_function}

Here we establish the objective for training the proposed style prompter. For a batch of inputs $\boldsymbol{X}$ randomly sampled from the training domains, we feed them into the image encoder (IE) to obtain image features \(\boldsymbol{Z}\). As mentioned earlier, the proposed style prompter maps these image features to the token embedding space, producing $N_S$ style samples $\boldsymbol{{S_o}}\in \mathbb{R}^{N_S\times D_t}$, where \(N_S\) (equal to $B$ for the basic style prompter and $B\times N$ for the Gaussian style prompter) denotes the number of style samples in the token embedding space, and \(D_t\) denotes the token embedding dimension.

\textbf{Open Domain Discrimination Loss.} For domain discrimination, we L2-normalize the generated style token embeddings, resulting in $N_S$ new data points $\boldsymbol{S}$ on the unit hypersphere. The samples from the same domain as the data point $\boldsymbol{s}_i$ are marked as positive samples, while those from different domains are marked as negative samples. The open domain discrimination loss for the data point $\boldsymbol{s}_i$ is defined as follows: 
\begin{equation}
\mathcal{L}_D^i = - \log{\frac{\sum_{\boldsymbol{s}^+_i \in \{D_i\},\boldsymbol{s}^+_i \neq \boldsymbol{s}_i} {\exp{( ({\boldsymbol{s}^\top_i}\boldsymbol{s}^+_i) / \tau)}}}{\sum_{j=1,j \neq i}^{N_S} {\exp{( ({\boldsymbol{s}^\top_i}\boldsymbol{s}_j) / \tau)}}}},
\label{L_domain}
\end{equation}  
where $\tau$ denotes a temperature parameter. The final domain discrimination loss is given by: $\mathcal{L}_{D} =  \frac{1}{N_S} \sum_{i=1}^{N_S} \mathcal{L}_{D}^{i}$.

\begin{figure}[t]
  \centering
  \includegraphics[width=1\linewidth]{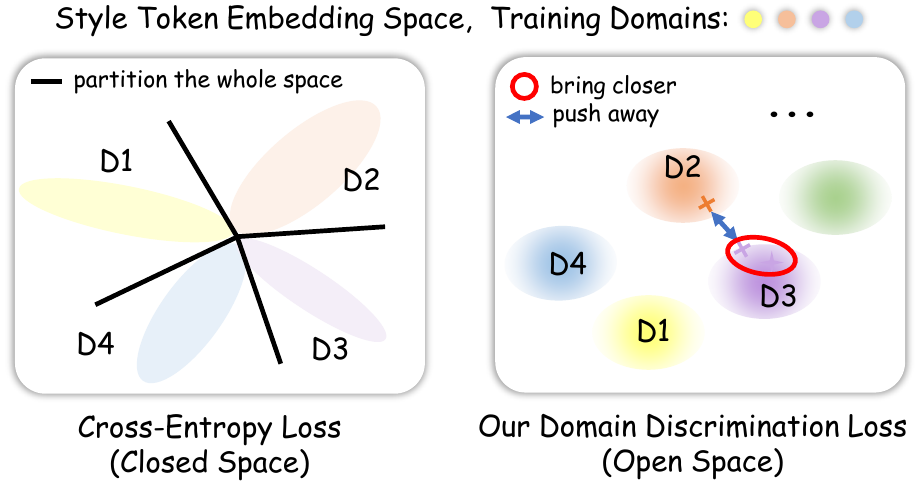}
  \caption{The common cross-entropy loss learns a partition of the whole embedding space based on training domains, cannot deal with unseen domains. The proposed domain discrimination loss shapes the embedding space by contrastive mechanism, leaving room for unknown style embeddings.}
  \label{openD_loss}
  \vskip -0.2in
\end{figure}

As illustrated in \myfigref{openD_loss}, compared to the commonly used cross-entropy classification loss, which assigns a class index in the range $[0, C-1]$ (where $C$ is the number of classes) as the target and learns a partition of the ``whole'' embedding space, our domain discrimination loss enables the trained style prompter to recognize open unknown domains. Specifically, our $\mathcal{L}_{D}$ loss brings the style embeddings of the same domain as close as possible and pushes the style embeddings of different domains far apart in the token embedding space. This approach does not learn a closed partition of the embedding space, leaving room for unknown style embeddings. Furthermore, as mentioned in the contrastive representation learning work by \cite{WuXYL18}, the apparent similarity in the data can naturally bring visually correlated classes closer than others, even if they are not labeled as the same class. Thus, our domain discrimination loss provides two crucial attributes for the trained style prompter: an open embedding space and a reasonable embedding distribution.

\textbf{Style Regularization Loss.} To further enhance the generalization of the trained style prompter, we introduce a hand-crafted prompt for regularization. Specifically, we select style words (including photo, art painting, cartoon, sketch, clipart, infograph, quickdraw, and product) based on common domains and incorporate them into the prompt template ``a [DOMAIN] style of a [CLASS].''. We then calculate a mean prompt $Z^{t, reg}_c$ for each class in the shared image-text feature space, representing a MEAN DOMAINS style of a [CLASS]. Finally, we use the hand-crafted prompt $Z^{t, reg}_c$ in the shared image text feature space to regularize the text features $Z^t$ during training. The loss function is as follows:
\begin{equation}
\mathcal{L}_{reg} = \frac{1}{B} \sum_{i=1}^{B} \Big(1 - CosSim(Z^t_{i, c}, Z^{t, reg}_{c})\Big),
\label{L_reg}
\end{equation}
where $CosSim(.,.)$ calculates the cosine similarity between two vectors.

The designed style regularization loss constrains the range of style embeddings generated by the style prompter, improving the stability of the training process and the generalization of the trained model. It also improves the contextual fusion of generated styles and category words. Subsequent experiments verify that this regularization benefits our method, leading to better results.

\textbf{Classification Loss.} To facilitate the current task, we also add the cross-entropy loss $\mathcal{L}_{CE}$, as in CLIP (contrastive language image pre-training \cite{RadfordKHRGASAM21}), for classification in the shared image-text feature space. This encourages the learned style embeddings and category words to complement each other, benefiting the task at hand.

Finally, we combine the domain discrimination loss $\mathcal{L}_{D}$ and regularization loss $\mathcal{L}_{reg}$ with the classification loss $\mathcal{L}_{CE}$, using weight parameters to balance their contributions. The final objective used for training is as follows:
\begin{equation}
\mathcal{L} =  w_d \times \mathcal{L}_{D} + w_{reg} \times \mathcal{L}_{reg} + \mathcal{L}_{CE}.
\label{L_final}
\end{equation}
The entire training framework is illustrated in \myfigref{framework_train}, where only the orange module, our style prompter, updates its parameters during training.

\subsection{Inference}
\label{Inference}
When a test image is received, we first obtain the image features with the image encoder, initialized with weights from the pre-trained vision-language model. Next, we extract its style (distribution) information as style priors (SP) using the trained style prompter, placing it in front of the candidate category words to construct the candidate texts (``SP [CLASS].''). We obtain candidate text features by sending the candidate texts to the text encoder, which is also initialized with weights from the pre-trained vision-language model. Finally, by matching the image features with the candidate text features, we predict the class corresponding to the text features with the highest similarity.

\begin{table*}[t]
  \caption{Results of our approach compared with different DG methods on four commonly used public datasets. ZS-CLIP (C) denotes zero-shot CLIP using ``[CLASS]” as its text prompt, and ZS-CLIP (PC) indicates zero-shot CLIP using “a photo of a [CLASS]” as its text prompt. $^\dagger$ denotes results from the DomainBed \cite{GulrajaniL21} paper.}
  \label{main_results}
  \centering
  \resizebox{0.77\linewidth}{!}{
  \begin{tabular}{lccccc}
    \toprule
    Method & PACS & VLCS & OfficeHome & \multicolumn{1}{l|}{DomainNet} & Average \\
    \midrule
    \multicolumn{6}{c}{ResNet-50 with pre-trained weights on ImageNet} \\
    \midrule
    DANN$^\dagger$ \cite{GaninUAGLLML16} & 83.6 & 78.6 & 65.9 & \multicolumn{1}{c|}{38.3} & 66.6 \\
    RSC$^\dagger$ \cite{HuangWXH20} & 85.2 & 77.1 & 65.5 & \multicolumn{1}{c|}{38.9} & 66.7 \\
    MLDG$^\dagger$ \cite{LiYSH18} & 84.9 & 77.2 & 66.8 & \multicolumn{1}{c|}{41.2} & 67.5 \\
    SagNet$^\dagger$ \cite{NamLPYY21} & 86.3 & 77.8 & 68.1 & \multicolumn{1}{c|}{40.3} & 68.1 \\
    SelfReg \cite{KimYPKL21} & 85.6 & 77.8 & 67.9 & \multicolumn{1}{c|}{41.5} & 68.2 \\
    GVRT \cite{MinPKPK22} & 85.1 & 79.0 & 70.1 & \multicolumn{1}{c|}{44.1} & 69.6 \\
    MIRO \cite{ChaLPC22} & 85.4 & 79.0 & 70.5 & \multicolumn{1}{c|}{44.3} & 69.8 \\
    \midrule
    \multicolumn{6}{c}{ResNet-50 with pre-trained weights from CLIP} \\
    \midrule
    \rowcolor{gray!7}
    ZS-CLIP (C) & 91.0 & 80.7 & 68.6 & \multicolumn{1}{c|}{46.2} & 71.6 \\
    \rowcolor{gray!7}
    ZS-CLIP (PC) & 91.0 & 81.4 & 72.0 & \multicolumn{1}{c|}{46.7} & 72.8 \\
    CAD \cite{RuanDM22} & 92.0 & 82.3 & 71.9 & \multicolumn{1}{c|}{48.8} & 73.8 \\
    PromptStyler \cite{ChoNKYK23}& 93.2 & 82.3 & 73.6 & \multicolumn{1}{c|}{49.5} & 74.7 \\
    \rowcolor{blue!5}
    Ours & \textbf{93.5} & \textbf{83.0} & \textbf{75.6} & \multicolumn{1}{c|}{\textbf{50.7}} & \textbf{75.7} \\
    \midrule
    \multicolumn{6}{c}{ViT-B/16 with pre-trained weights from CLIP} \\
    \midrule
    \rowcolor{gray!7}
    ZS-CLIP (C)  & 95.8 & 77.2 & 79.7 & \multicolumn{1}{c|}{57.7} & 77.6 \\
    \rowcolor{gray!7}
    ZS-CLIP (PC) & 96.1 & 82.1 & 82.4 & \multicolumn{1}{c|}{57.7} & 79.6 \\
    MIRO \cite{ChaLPC22} & 95.6 & 82.2 & 82.5 & \multicolumn{1}{c|}{54.0} & 78.6 \\
    PromptStyler \cite{ChoNKYK23} & \textbf{97.2} & \textbf{82.9} & 83.6 & \multicolumn{1}{c|}{59.4} & 80.8 \\
    \rowcolor{blue!5}
    Ours & \underline{97.1} & \underline{82.8} & \textbf{84.6} & \multicolumn{1}{c|}{\textbf{60.8}} & \textbf{81.3} \\
    \midrule
    \multicolumn{6}{c}{ViT-L/14 with pre-trained weights from CLIP} \\
    \midrule
    \rowcolor{gray!7}
    ZS-CLIP (C)  & 97.7 & 77.8 & 85.8 & \multicolumn{1}{c|}{63.2} & 81.1 \\
    \rowcolor{gray!7}
    ZS-CLIP (PC) & 98.4 & 80.7 & 86.6 & \multicolumn{1}{c|}{63.4} & 82.3 \\
    PromptStyler \cite{ChoNKYK23}& 98.6 & 82.4 & \textbf{89.1} & \multicolumn{1}{c|}{65.5} & 83.9 \\
    \rowcolor{blue!5}
    Ours & \textbf{98.8} & \textbf{83.2} & \textbf{89.1} & \multicolumn{1}{c|}{\textbf{66.2}} & \textbf{84.3} \\
    \bottomrule
  \end{tabular}}
  \vskip -0.1in
\end{table*}

\begin{figure*}[t]
  \centering
  \includegraphics[width=0.93\linewidth]{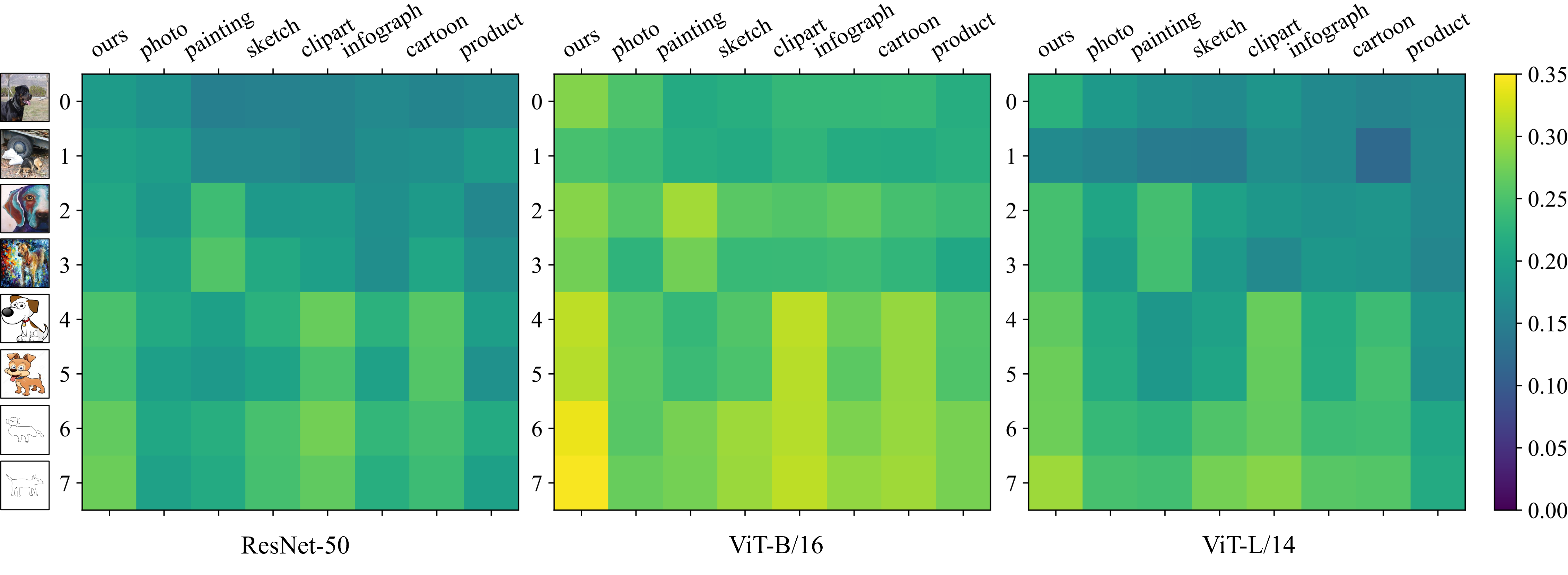}
  \vskip -0.1in
  \caption{Feature similarities between test images from unseen domains and texts generated with different style words. Compared to the unmatched style words, the learned styles and the artificially defined matching ones have higher similarities with the test image, indicating that our style prompter can correctly extract the style information of unknown images}
  \label{visualization}
  \vskip -0.1in
\end{figure*}

\section{Experiments}

\subsection{Evaluation Datasets}

We evaluate our method on four widely used multi-domain datasets: PACS \cite{LiYSH17}, VLCS \cite{FangXR13}, Office-Home \cite{VenkateswaraECP17}, and DomainNet \cite{PengBXHSW19}. PACS has seven categories (dog, elephant, giraffe, guitar, horse, house, person) across four domains (Photo, Art Painting, Cartoon, Sketch) with 9,991 images, offering a clean and balanced multi-domain setup. VLCS includes five classes with 10,729 images from four object-centric and scene-centric datasets (Caltech-101, PASCAL VOC2007, LabelMe, SUN09), each treated as a separate domain. Office-Home comprises 65 categories with 15,588 images across four domains (Art, Clipart, Product, Real-World), focusing on objects commonly found in office and home settings. DomainNet, a large-scale dataset, contains 345 classes and 586,575 images spanning 6 domains (clipart, infograph, painting, quickdraw, real, sketch).

\subsection{Implementation Details}
In our baseline, we directly use the pre-trained vision-language model (CLIP, Contrastive Language Image Pre-training) \cite{RadfordKHRGASAM21} to test different domains of the dataset and calculate the averaged classification accuracy. In our method, we follow the same leave-one-domain-out protocol as the compared DG methods, which first trains the model on multiple source domains and then tests it on the entire held-out domain. For data processing, we randomly divide each training domain into the training set (90\%) and validation set (10\%) and use the same data augmentation as CLIP. For the hyperparameters, we uniformly set the number of Monte Carlo samples to 40, the weight parameter $w_{D}$ of the domain discrimination loss to 0.1, and the weight parameter $w_{reg}$ of the regularization loss to 1 or 10 (1 for RN50 backbone, 10 for ViT backbone). For training, our style prompter only needs to train for three epochs in most cases, using SGD optimizer with momentum 0.9, weight decay $5\times 10^{-4}$, initial learning rate 0.002, and cosine annealing learning rate scheduler. We also use the warmup trick during the first epoch by fixing the learning rate to $1\times 10^{-5}$.

\subsection{Evaluations} 
\textbf{Main Results.} Our method significantly improves the model's domain generalization ability compared with the original CLIP, achieving state-of-the-art average performance across four public datasets, as shown in \mytabref{main_results}. Specifically, we investigate the performance of different models using ResNet-50, ViT-B/16, and ViT-L/14 as the image encoder, respectively. Compared with CLIP-based DG methods (CAD, MIRO, and PromptStyler), which only use the image encoder with pre-trained weights from CLIP and a closed set of classifiers specially trained on downstream tasks for inference, our method retains CLIP's classification capability, making the trained model more scalable, robust, and capable of handling recognition tasks in the open world. By extracting the style information from the test image and using it as prior knowledge to dynamically adapt the model, our method achieves excellent performance on difficult DG tasks, with particularly strong results on DomainNet.

\begin{table}[t]
\caption{Effects of different components on model performance. ``BSP'', ``GSP'' and ``SR'' denote ``Basic Style Prompter'', ``Gaussian Style Prompter'' and ``Style Regularization'', respectively. Baseline is zero-shot CLIP using ``[CLASS]” as its text prompt. Each component effectively enhances the baseline model's performance.}
\label{ablation_component}
\centering
\resizebox{0.95\linewidth}{!}{
\begin{tabular}{lcccc}
\toprule
Method     & PACS & VLCS & \multicolumn{1}{c|}{OfficeHome} & Average \\ \midrule
Baseline   & 91.0 & 80.7 & \multicolumn{1}{c|}{68.6}       & 80.1    \\
\rowcolor{blue!2}
+ BSP       & 91.6 & 81.3 & \multicolumn{1}{c|}{75.2}       & 82.7   \\
\rowcolor{blue!3}
+ GSP      & 92.3 & 81.9 & \multicolumn{1}{c|}{75.4}       & 83.2    \\
\rowcolor{blue!5}
+ GSP + SR & \textbf{93.5} & \textbf{83.0} & \multicolumn{1}{c|}{\textbf{75.6}}       & \textbf{84.0}    \\ 
\bottomrule
\end{tabular}}
\vskip -0.2in
\end{table}

\noindent \textbf{Component Effectiveness.} We investigate the impact of different components in our method on model performance through ablation studies on the PACS, VLCS, and OfficeHome datasets. The results in \mytabref{ablation_component} show that the proposed style priors are crucial for enhancing performance. The basic style prompter achieves a 2.6\% average improvement over the baseline model, while the Gaussian style prompter, which models style embeddings of input images as Gaussian distributions and samples from these distributions during training, improves performance by 3.1\%. These results validate the effectiveness of using style priors to prompt the model at test time for better domain generalization. Additionally, the Gaussian style prompter trained with our style regularization constraints, represented as ``a MEAN DOMAINS style of a [CLASS]. '' provides a 3.9\% improvement over the baseline model, highlighting the effectiveness of our style regularization loss in enhancing the prompter's ability to generalize across unseen domains.

\noindent \textbf{Visualization Analysis.} We conduct visualization analyses to assess the representation capabilities of our style prompter when combined with various backbones. Specifically, we investigate the style information extracted by models trained on the OfficeHome dataset when applied to unseen domains in PACS. To ensure disjoint training and test domains, we selected models trained on the Art, Clipart, and Product domains within the OfficeHome dataset, employing ResNet-50, ViT-B/16, and ViT-L/14 as image encoders, respectively. The test images are drawn from four domains in the PACS dataset: photo, art painting, cartoon, and sketch. Within the joint vision-language feature space, we compute the similarity between each test image and the textual descriptions generated based on both learned and artificially defined styles. As illustrated in \myfigref{visualization}, the learned styles and the artificially defined matching styles have higher similarities with test images compared to unmatched styles, demonstrating that our style prompter effectively extracts style information from test images in unseen domains.

\subsection{More Analyses} 

\noindent \textbf{StylePrompter vs. CoCoOp vs. Bayesian CoCoOp.} Compared to our StylePrompter, CoCoOp \cite{ZhouYL022} and Bayesian CoCoOp \cite{DerakhshaniSBCS23} also dynamically adapt the trained model using current input images at test time. CoCoOp employs a lightweight neural network to generate a content token embedding for each input image, which is then combined with learnable context prompts. Bayesian CoCoOp, building on CoCoOp, regularizes the prompt space from the Bayesian perspective by modeling it as a prior distribution. Both methods aim to improve the generalization of prompts on unseen classes by incorporating content information from the input image, without special consideration for domain shifts. As shown in \mytabref{style_advantage}, with the same training data, our StylePrompter has a greater ability to cope with distribution shifts by introducing style priors to the model at test time.

\begin{table}[t]
  \caption{Results of our approach compared with CoCoOp and Bayesian CoCoOp on cross-domain generalization. CLIP employs ``a photo of a [CLASS]'' as its text prompt. The proposed method has stable advantages on multiple datasets and different backbones.}
  \label{style_advantage}
  \centering
  \resizebox{\linewidth}{!}{
  \begin{tabular}{lcccc}
    \toprule
    Method & PACS & VLCS & \multicolumn{1}{c|}{OfficeHome} & Average \\
    \midrule
    \multicolumn{5}{c}{ResNet-50 with pre-trained weights from CLIP} \\
    \midrule
    CLIP & 91.0 & 81.4 & \multicolumn{1}{c|}{72.0} & 81.5 \\
    CoCoOp & 92.9 & 81.2 & \multicolumn{1}{c|}{74.7} & 82.9 \\
    Bayesian CoCoOp & 93.3 & 81.0 & \multicolumn{1}{c|}{75.2} & 83.2 \\
    \rowcolor{blue!5}
    Ours & \textbf{93.5} & \textbf{83.0} & \multicolumn{1}{c|}{\textbf{75.6}} & \textbf{84.0} \\
    \midrule
    \multicolumn{5}{c}{ViT-B/16 with pre-trained weights from CLIP} \\
    \midrule
    CLIP & 96.1 & 82.1 & \multicolumn{1}{c|}{82.4} & 86.9 \\
    CoCoOp & 96.8 & 82.2 & \multicolumn{1}{c|}{\textbf{84.9}} & 88.0 \\
    Bayesian CoCoOp & 96.6 & 80.8 & \multicolumn{1}{c|}{84.6} & 87.3 \\
    \rowcolor{blue!5}
    Ours & \textbf{97.1} & \textbf{82.8} & \multicolumn{1}{c|}{84.6} & \textbf{88.2} \\
    \midrule
    \multicolumn{5}{c}{ViT-L/14 with pre-trained weights from CLIP} \\
    \midrule
    CLIP & 98.4 & 80.7 & \multicolumn{1}{c|}{86.6} & 88.6 \\
    CoCoOp & 98.1 & 79.7 & \multicolumn{1}{c|}{89.0} & 88.9 \\
    Bayesian CoCoOp & 98.7 & 81.9 & \multicolumn{1}{c|}{\textbf{89.5}} & 90.0 \\
    \rowcolor{blue!5}
    Ours & \textbf{98.8} & \textbf{83.2} & \multicolumn{1}{c|}{89.1} & \textbf{90.4} \\
    \bottomrule
  \end{tabular}}
  \vskip -0.2in
\end{table}

\noindent \textbf{Potential across Categories and Domains.} To evaluate the potential of our proposed approach for zero-shot generalization—encompassing both cross-domain and cross-category settings—we maintain disjoint training and test domains, as well as disjoint training and test categories. Specifically, we utilize Art, Clipart, and Product from the OfficeHome dataset as training domains and evaluate performance on the four domains (Art Painting, Cartoon, Photo, and Sketch) in the PACS dataset. The experimental results presented in \mytabref{OfficeHome_to_PACS} demonstrate that our method preserves CLIP's zero-shot transfer capability while concurrently enhancing its cross-domain performance relative to the original CLIP.

\begin{table}[t]
  \caption{Zero-shot generalization. We train our model on the OfficeHome dataset and evaluate it on the PACS dataset, where the training and test data share different label and domain spaces. CLIP uses ``[CLASS]” as its text prompt.}
  \label{OfficeHome_to_PACS}
  \centering
  \resizebox{0.95\linewidth}{!}{
  \begin{tabular}{lccccc}
    \toprule
    Method & Art Painting & Cartoon & Photo & \multicolumn{1}{l|}{Sketch} & Average \\
    \midrule
    \multicolumn{6}{c}{ResNet-50 with pre-trained weights from CLIP} \\
    \midrule
    CLIP & 89.4 & 95.3 & 99.5 & \multicolumn{1}{c|}{79.7} & 91.0 \\
    \rowcolor{blue!5}
    Ours& \textbf{93.4} & \textbf{95.5} & \textbf{99.5} & \multicolumn{1}{c|}{\textbf{82.2}} & \textbf{92.6} \\
    \midrule
    \multicolumn{6}{c}{ViT-B/16 with pre-trained weights from CLIP} \\
    \midrule
    CLIP & 96.9 & 98.8 & 99.9 & \multicolumn{1}{c|}{87.7} & 95.8 \\
    \rowcolor{blue!5}
    Ours& \textbf{97.9} & \textbf{99.1} & \textbf{99.9} & \multicolumn{1}{c|}{\textbf{92.1}} & \textbf{97.3} \\
    \midrule
    \multicolumn{6}{c}{ViT-L/14 with pre-trained weights from CLIP} \\
    \midrule
    CLIP & 97.6 & 99.4 & 99.9 & \multicolumn{1}{c|}{93.7} & 97.7 \\
    \rowcolor{blue!5}
    Ours& \textbf{98.7} & \textbf{99.6} & \textbf{100} & \multicolumn{1}{c|}{\textbf{95.9}} & \textbf{98.6} \\
    \bottomrule
  \end{tabular}}
  \vskip -0.2in
\end{table}

\section{Conclusion}
Existing domain generalization methods that directly apply trained models to handle unseen domains without adaptation struggle to ensure effectiveness in facing any unseen domains. In this paper, we propose to dynamically adapt the trained model to the current test environment using style priors. Inference is performed offline through image-text matching without adapting the model parameters. In detail, we train a style prompter to extract the style information of the input image and encode it into a style token embedding, which is positioned before the class name as prior knowledge within the language modality to prompt the model. Our open partition of the style token embedding space, along with hand-crafted style regularization, enables the trained style prompter to effectively extract style information from samples in unseen domains.  Extensive experiments validate the effectiveness of our method, demonstrating state-of-the-art performance on four commonly used public datasets. Future work will explore leveraging the reasoning capability of large language models to assist in visual task learning, thereby further improving the domain generalization ability of the trained models.

\bibliography{aaai25}

\end{document}